\documentclass[lettersize,journal]{IEEEtran}
\usepackage{amsmath,amsfonts}
\usepackage{algorithmic}
\usepackage{algorithm}
\usepackage{array}
\usepackage[caption=false,font=normalsize,labelfont=sf,textfont=sf]{subfig}
\usepackage{textcomp}
\usepackage{stfloats}
\usepackage{url}
\usepackage{verbatim}
\usepackage{graphicx}
\usepackage{cite}
\usepackage{amsmath}

\usepackage{graphicx}
\graphicspath{{Figures/}}
\usepackage{xcolor}
\usepackage{float}
\usepackage{array}
\usepackage{multirow}
\usepackage{colortbl}
\usepackage{booktabs}

\makeatletter

\newcommand{\Rmnum}[1]{\expandafter\@slowromancap\romannumeral #1@}
\makeatother

\definecolor{mycolor}{gray}{0.9}

\begin{document}

\title{Monocular Depth Estimation From the Perspective of Feature Restoration: A Diffusion Enhanced Depth Restoration Approach }

\author{Huibin Bai\textsuperscript{*}, Shuai Li\textsuperscript{*},~\IEEEmembership{Senior Member,~IEEE}, Hanxiao Zhai,  Yanbo Gao\textsuperscript{\dag}, Chong Lv,  Yibo Wang, \\Haipeng Ping, Wei Hua, Xingyu Gao
	
\thanks{H. Bai and Y. Gao are with School of Software, Shandong University, Jinan 250100, China, and also with Shandong University-WeiHai Research Institute of Industrial Technology, Weihai 264209, China.
\par S. Li, H. Zhai and C. Lv are with School of Control Science and Engineering, Shandong University, Jinan, China. 
\par Y. Wang and H. Ping is with Shandong Institute of Information Technology Industry Development, Jinan, China. \par W. Hua is with Research Institute of Interdisciplinary Innovation, Zhejiang Lab, Hangzhou, China.   \par Xingyu Gao is with the Institute of Microelectronics, Chinese Academy of Sciences, Beijing 100029, China, and also with the University of Chinese Academy of Sciences, Beijing 100049, China. 
	
}

}



\maketitle

{ \begin{abstract}	
Monocular Depth Estimation (MDE) is a fundamental computer vision task with important applications in 3D vision. The current mainstream MDE methods {employ} an encoder-decoder architecture with multi-level/scale feature processing. However, the {limitations} of the current architecture and {the} effects of different-level features on the prediction accuracy {are not evaluated}. In this paper, we first investigate {the above problem} and show that there is still substantial potential in the current framework if  encoder features {can be improved}. Therefore, we propose to formulate the depth estimation problem from the feature restoration perspective, by treating {pretrained} encoder features as degraded features of an assumed ground truth feature that yields the ground truth depth map. Then an Invertible Transform-enhanced Indirect Diffusion (InvT-IndDiffusion) module is developed for feature restoration. {Due to the absence of} direct supervision on feature, only indirect supervision from the final sparse depth map is {used}. During the iterative {procedure of} diffusion, this results in {feature deviations} among steps. The proposed InvT-IndDiffusion solves this problem by using an invertible transform-based decoder under {the} bi-Lipschitz condition. Finally, a plug-and-play Auxiliary Viewpoint-based Low-level Feature Enhancement module (AV-LFE) is developed to enhance local details with auxiliary viewpoint when available. Experiments demonstrate that the proposed method achieves {better performance} than the state-of-the-art methods on various datasets. Specifically on the KITTI benchmark, compared with the baseline, the performance is improved by 4.09\% and 37.77\% under different training settings {in terms of} RMSE. Code is available at \url{https://github.com/whitehb1/IID-RDepth}.

\end{abstract} }


\section{Introduction}

Monocular Depth Estimation (MDE) is a fundamental task in the field of 3D vision. It is crucial for many visual applications such as {autonomous driving} and 3D reconstruction~\cite{10348603,9964065,10019290, 8374888,ling2021unsupervised,10261254,Bayesian,ke2024repurposing}. It is known that MDE is an ill-posed problem, since depth prediction is inherently a three-dimensional problem and the depth value cannot be accurately determined by a single image~\cite{9833523,10325550,LiftFormer2025lift}. With the development of neural networks and {their} powerful learning ability, visual clues, including perspective deformation, occlusion {relationships} and focal {lengths}~\cite{8764412, 10552238,10403933,9669049}, can be used to estimate the depth value. Deep {learning-based} MDE, usually with an encoder-decoder architecture, has achieved significant improvements. However, with its ill-posed nature, whether MDE can be further improved remains {an open question}. In other words, whether effective features can be learned to predict a high-quality depth map {is still a problem} under the current network architectures. Moreover, how the different levels of encoder features affect the final depth prediction is not thoroughly investigated yet.

On the other hand, the diffusion model~\cite{dhariwal2021diffusion,ho2020denoising,Lugmayr_2022_CVPR,rombach2022high} has achieved excellent results in the image generation task, and many works have applied the diffusion model to dense prediction tasks such as semantic segmentation~\cite{wu2023diffumask,peng2023diffusion,toker2024satsynth}, {super-resolution}~\cite{shang2024resdiff,wang2024sinsr,li2022srdiff}, and {deblurring}~\cite{wu2024id,ren2023multiscale}. Recent studies {have also} investigated using the denoising diffusion model for the MDE task. Some works~\cite{patni2024ecodepth,lavreniuk2023evp} employ pre-trained Stable Diffusion~\cite{rombach2022high} (SD) as a depth feature generator, where general encoders such as VAE~\cite{kingma2013auto} are used as the backbone and CLIP features or other semantic features are used as conditional maps {for predicting} depth maps.  Other studies~\cite{Ke_2024_CVPR} {construct} virtual color image-dense depth  {map} pairs, and use images as conditional maps to train a denoising diffusion model. These works demonstrate the potential of using the diffusion model in depth estimation.

\begin{figure*}[t]
	\centering
    \vspace{0pt}
	\includegraphics[width=1\linewidth]{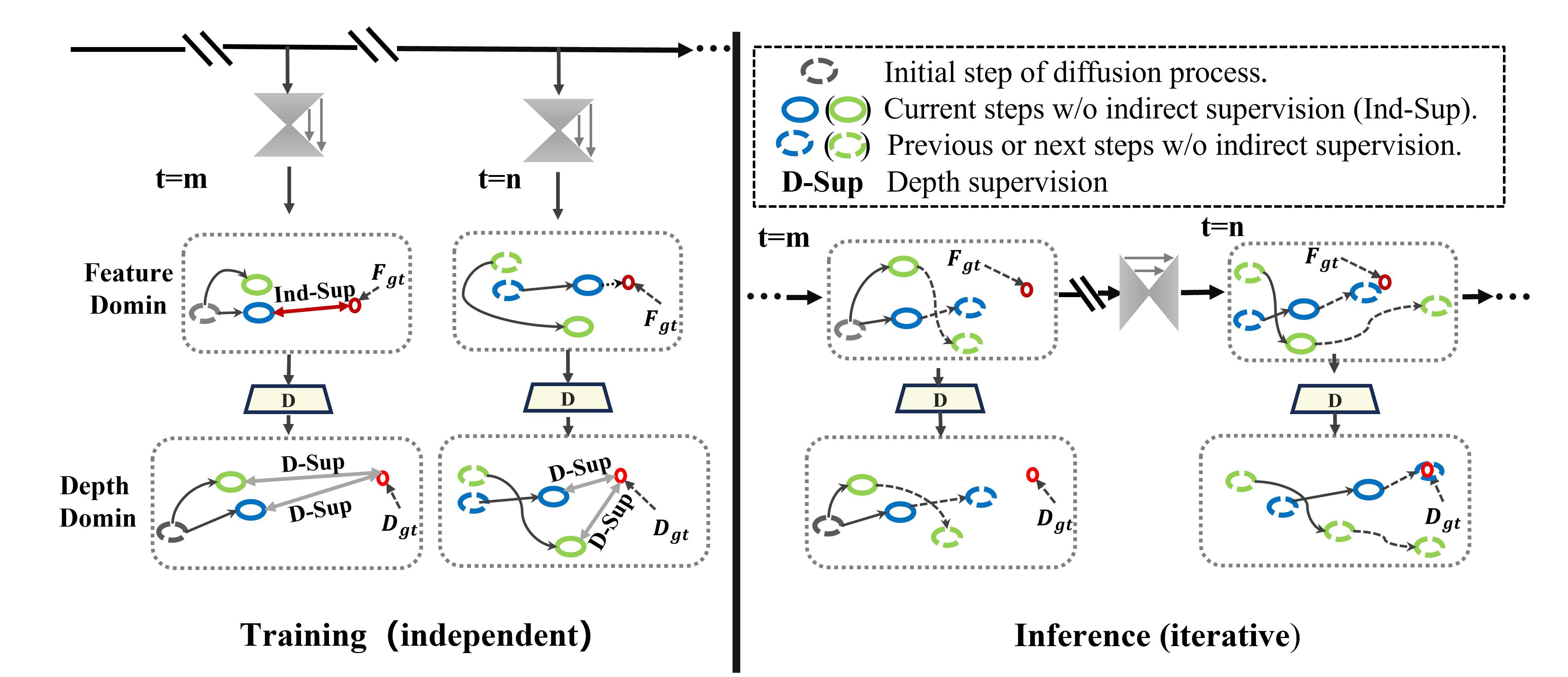}
     \vspace{-10pt}
	\caption{Illustration of the feature deviation problem in the diffusion process due to the indirect supervision and iterative inference. It can be mitigated with the proposed InvT-IndDiffusion based on the invertible decoder under bi-Lipschitz condition. }
	
	\vspace{-0.3cm}
	\label{fig:idea}
\end{figure*}

However, existing methods still face several challenges.  In real-world scenarios, the ground truth depth map is usually obtained from Lidar point clouds and in a sparse form. This complicates the diffusion process when adding noise to the ground truth in the forward propagation, thus making the reverse inference less effective. Additionally, the aforementioned method based on the {latent-diffusion} trains the model with supervision on the final depth prediction loss, which ignores the iterative property of the diffusion model. The lack of supervision on the latent features, which are iteratively used in the diffusion, may lead to latent feature deviations in iterations, compromising the final result when performing inference with multiple steps. As illustrated in Fig. \ref{fig:idea}, the indirect supervision of the latent diffusion model through depth maps reduces the overall loss of the model (as the gray dot {approaches} the ground truth depth map in the $m$-th step in Fig. \ref{fig:idea}), but it does not constrain the diffusion model in the latent feature space. Consequently, {with each step trained independently}, the feature cannot be consistently optimized towards the assumed ground truth feature (corresponding to the ground truth depth map) in the feature space (as the optimization in the $n$-th step in Fig. \ref{fig:idea}). In this case, in the inference, the feature may deviate from the assumed ground truth feature in the iterative generation process, leading to deviated depth prediction (as the inference in the $n$-th step in Fig. \ref{fig:idea}).  

To address the above problems, the effect of different-level encoder features on the final performance is first investigated. It is observed that better encoder features can significantly improve the depth estimation performance, as illustrated in the following Section Motivation. And more importantly, higher-level features affect the result more substantially than the lower-level ones.  Therefore, this paper formulates the MDE as a feature restoration problem and proposes an Invertible transform enhanced Indirect Diffusion for Restored Depth Estimation (IID-RDepth) framework. It restores the high-level semantic features through an Invertible Transform-enhanced Indirect Diffusion (InvT-IndDiffusion), which is developed to solve the problem of feature deviation in inference iterations with indirect supervision on the latent diffusion. For low-level features, an Auxiliary Viewpoint based Low-level Feature Enhancement module (AV-LFE) is developed to fully explore the detail information when an auxiliary viewpoint is available. 

To address the above problems, the effect of different-level encoder features on the final performance is first investigated. It is observed that better encoder features can significantly improve the depth estimation performance, as illustrated in {Section Motivation below}. And more importantly, higher-level features affect the result more substantially than the lower-level ones.  Therefore, this paper formulates the MDE as a feature restoration problem and proposes an {Invertible Transform-enhanced} Indirect Diffusion for Restored Depth Estimation (IID-RDepth) framework. It restores the high-level semantic features through an Invertible Transform-enhanced Indirect Diffusion (InvT-IndDiffusion), which is developed to solve the problem of feature deviation in inference iterations with indirect supervision on the latent diffusion. For low-level features, an Auxiliary Viewpoint{-based} Low-level Feature Enhancement module (AV-LFE) is developed to fully explore the {detail} information when an auxiliary viewpoint is available. 

The contributions of this paper are summarized as follows.

\begin{itemize}
	 \item{The effects of different-level encoder features on depth prediction are thoroughly investigated. It is observed that the higher-level features can significantly improve the final performance {and their improvement is more significant} than that of lower-level features.}
	  \item{We propose an IID-RDepth framework, a novel approach leveraging diffusion models for depth estimation. It formulates depth estimation from the perspective of restoring the encoder features to {alleviate their} degradation and performs restoration through a latent diffusion model.}
	   \item{A novel Invertible Transform-enhanced Indirect Diffusion (InvT-IndDiffusion) is developed for feature restoration. It is designed for latent diffusion without {ground truth} features, {relying only on} indirect target loss. The invertible transform is used to construct a decoder, to enforce indirect supervision on {the} final target{,} {ensuring} {undeviating} latent feature optimization in inference iterations based on the bi-Lipschitz condition. {undeviating} latent feature optimization in inference iterations based on the bi-Lipschitz condition.}
	    \item{A plug-and-play AV-LFE module is {developed} to fully {exploit} {the auxiliary view} when available without affecting the MDE performance.}	
\end{itemize}
Extensive experiments demonstrate that the proposed IID-RDepth can achieve better performance than the state-of-the-art methods. {An} Ablation study is also conducted to verify the effectiveness of each module.

\section{Related Work}

\subsection{Depth Estimation}

Monocular depth estimation is an important 3D perception task. Existing methods explore an encoder-decoder architecture to extract multi-level features and {fuse} them in a U-Net style to predict the depth  as in~\cite{agarwal2023attention,yuan2022new,9669049}.
Recent methods {have explored} more prior knowledge on depth estimation into the neural network design. In~\cite{yang2023gedepth}, GEDepth proposed to use a ground embedding module to generate ground depth and fuse it with residual depth via a ground attention module. Wu et al.~\cite{9669049} introduced the side prediction aggregation module and spatial refinement loss to enhance depth estimation by leveraging spatial priors. It generates multi-scale depth maps during decoding and uses adversarial learning for supervision, improving the spatial consistency. In~\cite{wang2022probabilistic}, perspective geometry and object-level prior information {were} used to construct a geometric relation graph for depth propagation in order to correct the depth estimation results. 
Yang et al.~\cite{Bayesian} proposed a Bayesian DeNet to leverage temporal information in multiple frames. The network predicts depth maps with uncertainties, and then applies the Bayesian inference based on the camera poses to refine depth estimation. 
Newcrf~\cite{yuan2022new} and Va-depth~\cite{liu2023va} use fully-connected conditional random fields and first-order variational constraints, respectively, to capture relationships between image features.
ADPDepth~\cite{10243088} introduced a geometric consistency prior-based depth-pose consistency loss to eliminate the scale ambiguity and parallel convolutional branches to capture inter-channel correlations and extract multi-scale features.
In~\cite{liu2023single}, a multivariate Gaussian distribution was used to formulate scene depth and the covariance among pixels is learned for regulating the depth generation.

On the other hand, there are also methods focusing on developing new ways {for predicting depth values}. {Li et al.~\cite{9852314} formulated the task as a frequency-domain enhancement problem and proposed a frequency-based recurrent depth coefficient refinement scheme. It uses an RNN architecture to progressively refine depth coefficients in the frequency domain, improving both global accuracy and local detail preservation.} {Cao et al.~\cite{8764412} divided the depth range into several discrete bins and predicted the pixel-wise probability distribution across these bins, thereby modeling the depth estimation task as a classification problem. Subsequently, Adabin~\cite{bhat2021adabins} formulated the MDE task as a classification-regression task, producing depth values by predicting adaptive bin centers and associated probabilities.} {HA-Bins~\cite{10325550} proposed a hierarchical framework with adaptive bins to enhance the robustness and generalization of network predictions. It progressively refines the representation of bins and employs multi-scale supervision to improve prediction consistency.}
BinsFormer~\cite{li2024binsformer} proposed a bin embedding to improve the interaction between features and bin centers. 
IEBins~\cite{shao2024iebins} further proposed iterative elastic bins for refining depth estimates through iteration, while Localbins~\cite{bhat2022localbins} adjusted bin centers to align with local pixel areas. Pixlefomer~\cite{agarwal2023attention} proposed an attention-based U-net where attention is used to refine pixel queries from the decoder feature maps by cross-attend the higher resolution encoder features for depth prediction. Our paper formulates depth prediction as a feature restoration problem to restore an assumed ground truth encoder feature for the depth map.


\subsection{Diffusion-based Depth Estimation}

Diffusion models have achieved great success in generative tasks, and many works have extended diffusion models to various fields,  such as image enhancement{~\cite{jiang2023low}}, semantic segmentation~{\cite{wu2023diffumask}} and depth estimation~{\cite{ke2024repurposing,wei2024d}}. Different diffusion architectures have been developed including the  Denoising Diffusion Probabilistic Models (DDPM)~\cite{ho2020denoising}, Denoising Diffusion Implicit Models (DDIM)~\cite{song2020denoising}, Latent Diffusion Models (LDM)~\cite{dhariwal2021diffusion}, and Residual Denoising Diffusion Models (RDDM)~\cite{liu2024residual}. In depth estimation, due to the sparse nature of the LiDAR-based ground truth, it is difficult to directly train a diffusion model by propagating the sparse ground truth to noise according to a noise schedule and performing the corresponding reverse inference process. In~\cite{Ke_2024_CVPR}, virtual depth-color image pairs are generated for diffusion training, using noisy depth images as input and color images as condition.
The other methods usually take the encoder feature from a depth estimation network as input and add noise, then perform the reverse training using the final sparse depth map as supervision. In~\cite{ji2023ddp}, DDP proposed to use the diffusion model to generate the depth map from noise with features extracted from the input image as both input (adding noise) and condition. EVP~\cite{lavreniuk2023evp} proposed to use stable diffusion for depth estimation and the text information is extracted to {serve} as extra {conditions}. ECoDepth~\cite{patni2024ecodepth} extracts a detailed image embedding with ViT as the conditional map for the diffusion model, providing more semantic information for depth estimation. 
However, since the denoised features {are} not directly supervised and can be optimized in different directions at different iterations, such indirect optimization may result in feature deviations in the final iterative inference. This paper investigates this problem and proposes an invertible transform enhanced indirect diffusion to solve it.

\section{Motivation}
MDE is an ill-posed problem, {which uses} a single image to reconstruct the depth of a scene. Moreover, depth {maps} usually contain sharp edges with values distributed in a large range. Therefore, whether effective features can be learned and efficiently decoded to generate a high-quality depth map is still a problem under the current network architectures. Especially under the UNet-like architecture, the effect of different levels of features on the final depth prediction {is} not fully investigated yet. In this paper, the above problems are first studied to shed light on the further development of MDE.

\begin{figure}[t]
	\centering
    \vspace{0pt}
	\includegraphics[width=1\linewidth]{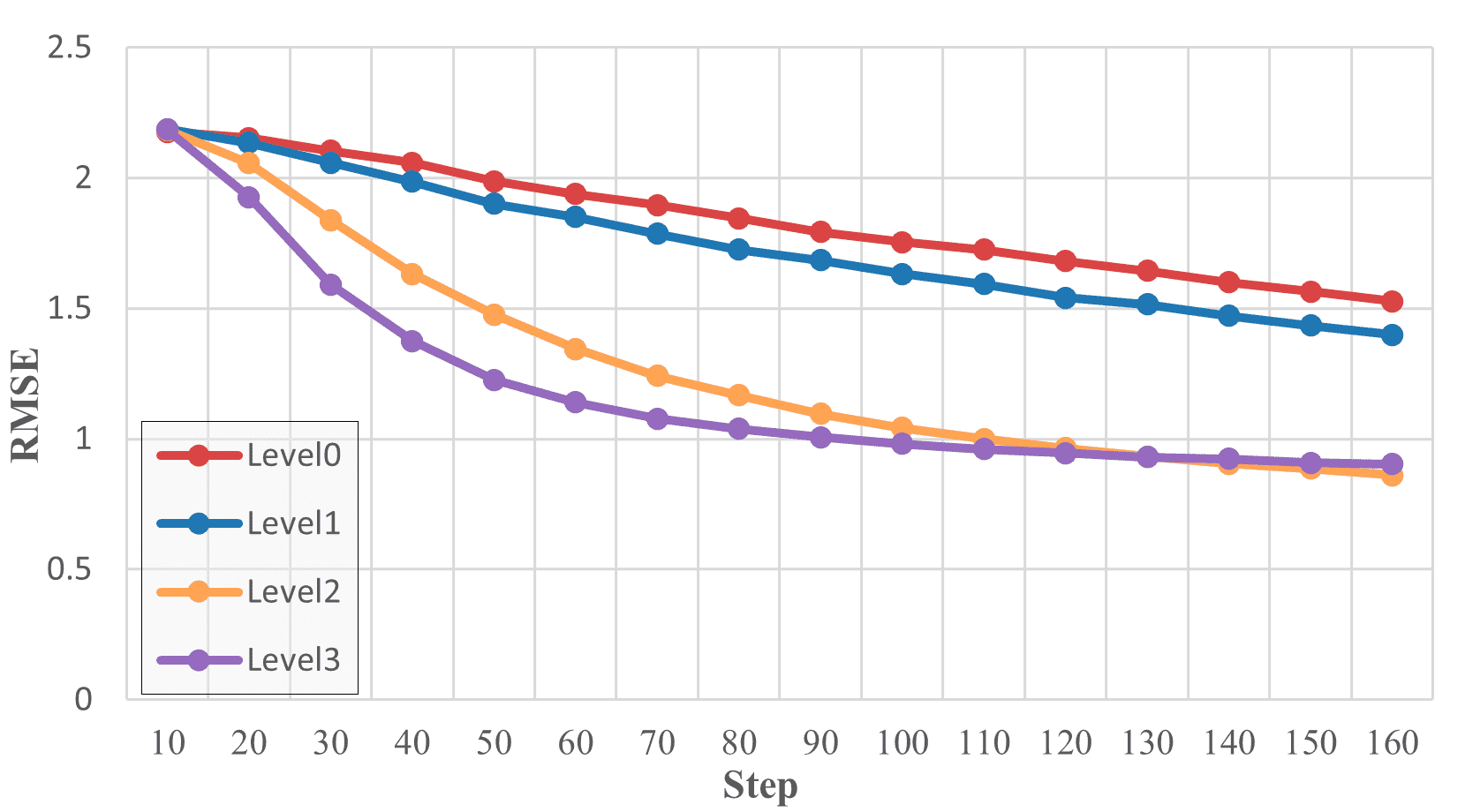}
     \vspace{-10pt}
	\caption{Effects of different levels of features on the final depth prediction performance in terms of RMSE (the smaller the better). }
	
	\vspace{-0.3cm}
	\label{fig:RMSE}
\end{figure}

{An existing MDE network, the PixelFormer~\cite{agarwal2023attention} to be specific, is used as the baseline for the experiment. It takes a Transformer-based U-Net architecture, with a four-stage feature processing. Down/up-sampling is used after each stage in the encoder and corresponding decoder to facilitate multi-scale processing, respectively. Accordingly, hierarchical (different-level/scale) features can be obtained from the different stages in the encoder. Pre-training is {adopted} with frozen network parameters to illustrate the effect of different encoders. Detailed settings are described in the Experimental setting. To investigate the impact of features from different levels on the final depth estimation, the hierarchical features obtained from the different-scale blocks in the encoder are optimized individually using a per-image optimization. The results are shown in Fig.~\ref{fig:RMSE} along the optimization steps, and several critical observations can be summarized below.}

{

\begin{itemize}
	 \item{{Performance is improved at all feature levels} even with a few optimization steps, indicating that the current network architecture can predict {more accurate depth maps} with improved features. }

\item{The performance can be significantly improved if features are optimized well with multiple steps. However, the converged performance {does not achieve } a very small error, also indicates that new architectures are {worthy of investigation} to reduce the depth estimation errors.}

\item{More importantly, {surprisingly} the higher-level features are more easily optimized and provide much {greater improvements} than the lower-level features, although depth estimation is a pixel-level task. This indicates the importance of the higher-level features.}
\end{itemize}

}

{Based on the above observations,} this paper proposes a high-level feature enhancement network for MDE. Moreover, considering that better higher-level features can be easily optimized within only a few steps as shown in Fig.~\ref{fig:RMSE}, in this paper, we treat the high-level feature enhancement as a restoring task by restoring a high-quality high-level feature from an initial degraded one. Given the nature that diffusion is used to generate better features by removing noise, diffusion, an invertible {transform-based} indirect diffusion model (presented later) to be specific, is used to perform the restoration. Since we solve the depth estimation task from the perspective of feature restoration with diffusion, the proposed method is {termed} Invertible transform enhanced Indirect Diffusion for Restored Depth Estimation (IID-RDepth).

\begin{figure*}[t]
	\centering
    \vspace{0pt}
	\includegraphics[width=0.9\linewidth]{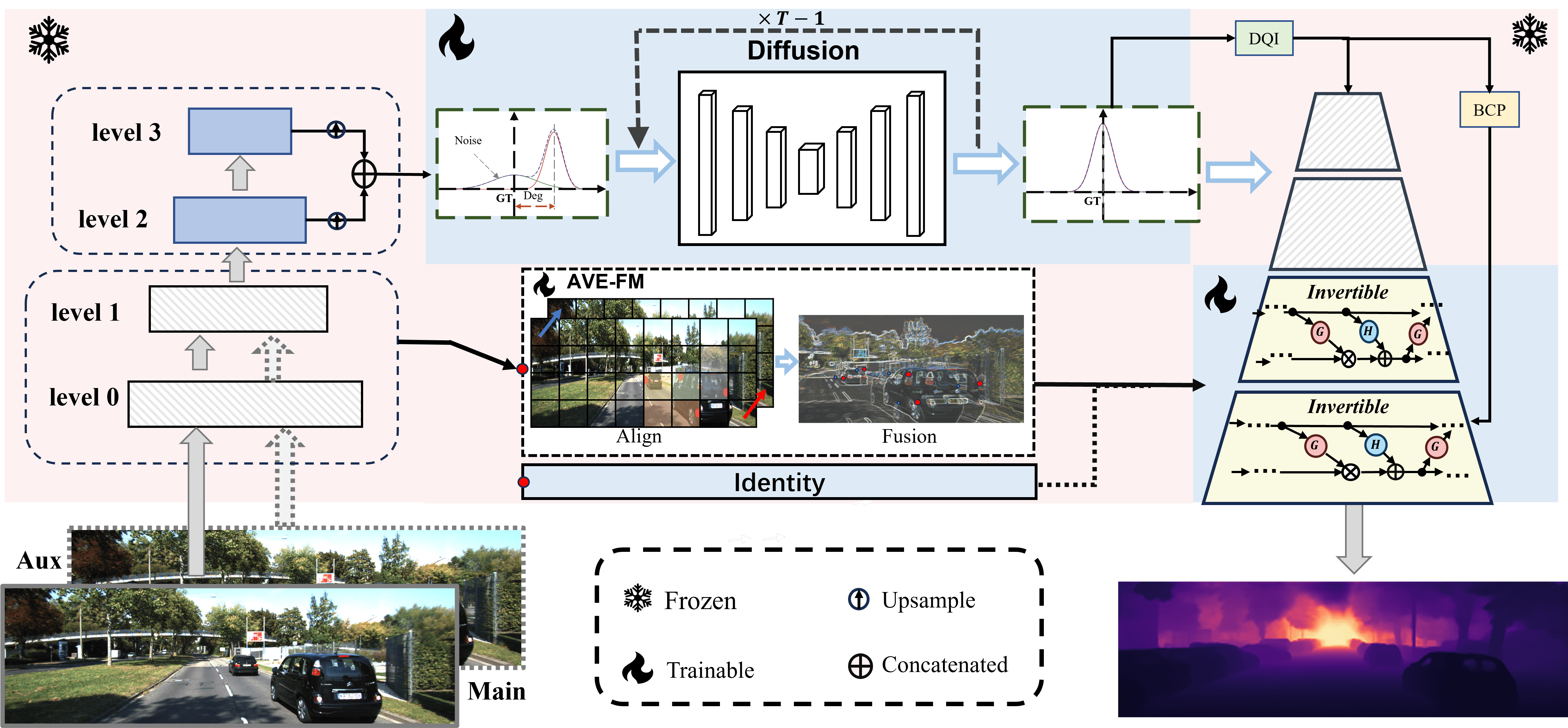}
    \vspace{-6pt}
	\caption{ Overview of the proposed IID-RDepth. The input image is first processed through a frozen encoder to extract the multi-level features. Then the high-level semantic features are combined as mutual conditional maps and fed into the InvT-IndDiffusion for restoration. For low-level detail features, AV-LFE is used as a plug-and-play { module} to enhance the identity connection when  auxiliary viewpoint is available.}
	\vspace{-0.1cm}
	\label{fig:frame}
\end{figure*}
\section{Proposed Method}

\subsection{Overview}

In this section, we present the proposed Invertible transform enhanced Indirect Diffusion for Restored Depth Estimation (IID-RDepth). The overall framework is shown in Fig.~\ref{fig:frame}.
{ It takes a U-Net architecture with {an} encoder and a decoder where each includes four stages/levels of feature processing. Down/up-sampling is applied sequentially after each stage in the encoder and corresponding decoder. The encoder employs Swin-Transformer blocks, which are pre-trained on depth estimation and fixed for feature extraction. The proposed InvT-IndDiffusion is used as the decoder for depth prediction, which contains a denoising diffusion network for restoration of high-level features and an invertible transform enhanced decoder. The high-level features from the encoder are concatenated and fed into our InvT-IndDiffusion as multi-scale mutual condition maps. {Adabin-based} depth prediction with the Decoder Query Initialiser (DQI) and the Bin Center Predictor (BCP) (detailed in the supplementary material) is used as in~\cite{agarwal2023attention}, which are also pre-trained and frozen by considering the denoising process as feature restoration. The low-level decoder is designed using the coupling layer based invertible structure.} In order to further improve the { low-level} features, an Auxiliary Viewpoint based Low-level Feature Enhancement module (AV-LFE) is proposed and used as a plug-and-play module to explore auxiliary {viewpoint} when available. {In the following, the proposed InvT-IndDiffusion module is first formulated in subsection \Rmnum{4}.B, then the overall IID-RDepth is presented in subsection \Rmnum{4}.C and the AV-LFE module is explained in subsection \Rmnum{4}.D.}

\subsection{Invertible Transform enhanced Indirect Diffusion (InvT-IndDiffusion)}
\label{bbb}
From the perspective of restoration for depth estimation, denoising diffusion models can be used to directly improve the predicted depth map or intermediate features. However, for supervised MDE, the ground truth depth maps are often derived from Lidar point clouds and in sparse representations. This makes it unsuitable for diffusion, which trains the model by adding noise to the input and then denoising it based on the spatial characteristics. On the other hand, latent diffusion, as in Stable Diffusion~\cite{rombach2022high}, has been widely studied for feature enhancement, where latent features added with noise are iteratively restored. However, there are no ground truth features that are directly mapped to the depth map{,} and simply supervising the diffusion model with the input {features} cannot provide good depth prediction {results}.
In this paper, we propose an {Invertible Transform-enhanced} Indirect Diffusion (InvT-IndDiffusion) to facilitate the latent feature diffusion with indirect optimization from the depth prediction. This process can be represented as

\begin{equation} 
	\label{direct_pred}
	D = f_m(f_{ID}(F_{in})), 
\end{equation}
where $D$ and $F_{in}$ represent the predicted depth map and the input {features}, respectively. $f_m$ and $f_{ID}$ represent the decoder function decoding the feature to predict {the} depth map, and the indirect diffusion function, respectively.

The key problem in such indirect diffusion is the misalignment between the optimization trajectories of feature and predicted depth. {Note that diffusion models are trained and inferred iteratively, with each iteration's output serving as input for the next.} Assume there {exists} a ground truth feature $F_{gt}$ that produces {a} ground truth depth map. In each diffusion step, the feature is supposed to be optimized towards $F_{gt}$, {thereby ensuring that} the result produced in each step can be further optimized in {subsequent} steps without deviating from $F_{gt}$. However, in indirect diffusion, {the optimization process} is supervised by the depth map, and the optimization at each step may deviate from $F_{gt}$ at different steps, {which renders the iterative inference ineffective during testing}.

To solve the above problem in indirect diffusion, the decoder function $f_m$ needs to satisfy: 
\begin{equation} 
	\small
	\label{argmin}
	\underset{F_n}{argmin} \left\|f_{m}(F_{n})-f_{m}(F_{g t})\right\| \xrightarrow{\text {yields}} \underset{F_n}{argmin} \left\|F_{n}-F_{g t}\right\|,
\end{equation}
where $F_{gt}$ represents the assumed ground truth feature producing the ground truth depth map in our restoration perspective. and $F_n$ represents the noisy input feature or the intermediate feature generated in the different steps of indirect diffusion. $|| \cdot ||$ is the norm to measure their distance. Accordingly, $f_m(F_{gt} )$ corresponds to the ground truth depth and $f_m(F_n)$ is the predicted depth. In other words, the reduction of the distance between depth values leads to the reduction of the distance between features. This can be realized by a distance-preserving mapping $||f_m (F_n )-f_m (F_{gt} )||=||F_n-F_{gt} ||$ or a relaxed version where the distance after the mapping is positively correlated. However, it is difficult to design a neural layer that fulfills the above strict constraint while keeping the strong nonlinear learning capability. To mitigate this, the  constraint is relaxed by utilizing the Lipschitz theorem, defined as follows:

where $F_{gt}$ represents the assumed ground truth feature producing the ground truth depth map in our restoration perspective. {And} $F_n$ represents the noisy input feature or the intermediate feature generated in different steps of indirect diffusion. $|| \cdot ||$ is {a} norm {used} to measure their distance. Accordingly, $f_m(F_{gt} )$ corresponds to the ground truth depth and $f_m(F_n)$ is the predicted depth. In other words, {the reduction of} the distance between depth values leads to {the reduction of} the distance between features. This can be realized by a distance-preserving mapping $||f_m (F_n )-f_m (F_{gt} )||=||F_n-F_{gt} ||$ or a relaxed version where the distance after {mapping} is positively correlated. However, it is difficult to design a neural layer that fulfills the above strict constraint while {keeping} {the } strong nonlinear learning {capability}. To mitigate this, {the constraint} is relaxed by {leveraging} the Lipschitz theorem, defined as follows:
\begin{equation} 
	\label{def_Lipschitz}
	||f(x_1)-f(x_2)||\le K||x_1-x_2||.
\end{equation}

With a Lipschitz continuous function $f$, the distance between the outputs of the function is always smaller than {or equal to} the distance between the original variable  multiplied {by} a real constant $K \ge  0$. By applying the above Lipschitz theorem to Eq.~\ref{def_Lipschitz}, it becomes:

\begin{equation} 
	\label{Our_Lipschitz}
	||F_n-F_{gt}||\ge\frac{1}{K}||f_m(F_n)-f_m(F_{gt})||.
\end{equation}
Assume the function $f_m$ is invertible, and satisfy bi-Lipschitz condition, the above Eq.~\ref{Our_Lipschitz} can be further constrained to:
\begin{equation} 
	\begin{aligned}
	\label{bi_Lipschitz}
	\frac{1}{K}||f_m(F_n)-f_m(F_{gt})|| \le||F_n-F_{gt}||, \\
	||F_n-F_{gt}|\le L||f_m(F_n)-f_m(F_{gt})||.
\end{aligned}
\end{equation}
In this way, as long as the decoder function is invertible and bi-Lipschitz continuous, the feature can be optimized towards the assumed groundtruth feature by minimizing the depth map loss. Such functions can be realized with the invertible residual network as in~\cite{dinh2014nice} or any invertible neural layer with {Lipschitz constraint} such as the weight clipping in~\cite{Arjovsky2017WassersteinGA}. Therefore, in this paper, we propose to use invertible neural layers to {construct} an invertible decoder module.

Specifically, the affine Coupling layer is used to construct the invertible decoder. Each coupling layer divides the input feature $F_{1:S}^{in}$ into two parts with a position $s < S$. One part is directly connected to the output as the key{,} while the other part is processed with {an} affine mapping where the affine factors are generated {from} the key through a neural network. This process is alternately performed within the coupling layer:
\begin{equation}
\begin{aligned}  
	\label{cp}
	F_{1:S}^{out} = F_{1:s}^{in}\odot exp(\sigma_c(g_2(F_{s+1:S}^{in})) )+ h_2(F_{s+1:S}^{in}), \\
	F_{s+1:S}^{out}=F_{s+1:S}^{in}\odot exp(\sigma_c(g_1(F_{1:s}^{out})) )+ h_1(F_{1:s}^{out}),
\end{aligned}
\end{equation}
where $\odot$ is the Hadamard product, {$\exp(\cdot)$ and $\sigma_c(\cdot)$ denote the exponential and sigmoid functions, respectively.} $g(\cdot)$ and $h(\cdot)$ represent the networks used to generate the affine factors.

In this paper, three coupling layers are used to construct the invertible decoder module. Together with the diffusion model, {it formulates} the Invertible Transform enhanced Indirect Diffusion (InvT-IndDiffusion). The proposed InvT-IndDiffusion can be {applied not only to} depth estimation but also to other latent diffusion models indirectly supervised by the final task.

\subsection{InvT-IndDiffusion based Feature Restoration for Depth Estimation}

As mentioned in {the} Motivation, optimization {of} the high-level features can bring {a} large performance improvement. Our goal is to design a denoising diffusion process for high-level semantic features to formulate a feature restoration process. The two high-level features are used as the {inputs} to InvT-IndDiffusion.

In the traditional DDPM, the inputs are progressively transformed {into} random noise in the forward process and {start} the inference from the random noise. However, for depth estimation, the task is to restore a specific target and {consider} the raw features contain all the encoder information essential to the final depth estimation, noise is progressively added according to a noise schedule while the {features} is kept the same without scaling. The forward process is defined as:
\begin{equation}
\begin{aligned} 
	\label{add_noise}
	F_t = &F_{t-1}+\alpha_t\varepsilon, \\     
	    =  &F_{t-2} + (\sqrt{\alpha_{t-1}^2+\alpha_t^2})\varepsilon\\ 
	    = &\cdots\\
	    = &F_{gt}+F_{deg}+\bar{\alpha _t}\varepsilon ,
\end{aligned}
\end{equation}
where $F_{gt}$ and $F_{deg}$ are the assumed ground truth {features} to be restored and the noise {features} contained in the input encoder {features}. $\varepsilon\sim\mathcal{N}(0,I)$ {is} the added noise from a Gaussian distribution, $\bar{\alpha}_t=\sqrt{ {\textstyle \sum_{i=1}^{t}\alpha _i^2} } $with $\bar{\alpha}_T = 1$ at the maximum forward step $T$, and the corresponding coefficient $\alpha_t$ is the noise schedule. By introducing noise, diverse degradation features can be generated and the structure of the original features can be partially disrupted, enabling them to escape potential local optima. These features can then be restored through the reverse inference.

When recovering the assumed ground truth features from the input noisy features, the features of the different layers are taken as condition maps to each other to take advantage of the multi-scale information. Pixel shuffle is employed to upsample the multi-scale features to {the} same scale, which are then added with noise and used as mutual condition {maps}  $(C_{mul})$ to the InvT-IndDiffusion network. Specifically, the diffusion model works as a restoration network $R_\theta (F_t,t,C_{mul} )$  to predict the degradation and added noise $F_{deg}$ and $\epsilon_t$.Then{,} from Eq.~\ref{add_noise}, the restored feature at each step can be obtained by $F_0^\theta=F_t-F_{deg}^\theta-(\bar{\alpha}_t -\bar{\alpha}_{t-1} ) \epsilon_t^\theta$, where $F_{deg}^\theta$ and $\epsilon_t^\theta$ are the predicted degradation and noise by the restoration network. Accordingly, the reverse probability can be expressed as: 
\begin{equation} 
	\label{reverse_probability}
	p_\theta(F_{t-1}|F_t):=q_\theta(F_{t-1}|F_t,F_0^\theta,F_{deg}^\theta,C_{mul}),
\end{equation}
where $q_\theta(\cdot)$ severs the transfer probability from $F_t$ to $F_{t-1}$. The restoration to the next step $t-1$ employs a deterministic sampling process similar to RDDM~\cite{liu2024residual}, which can be expressed by:
\begin{equation} 
	\label{sampling_process}
	F_{t-1} = F_t-F_{deg}^\theta-(\bar{\alpha}_t -\bar{\alpha}_{t-1} ) \epsilon_t^\theta.
\end{equation}
The final restoration feature can be obtained by iteratively performing this reverse inference. 

The SiLog loss proposed in~\cite{eigen2014depth} is adopted for supervision on the predicted depth map as:
\begin{equation} 
	\label{ind_sup}
	L(\theta) = E[||D_{gt}-D_{pre}||],
\end{equation}
where $D_{gt}$  and $D_{pre}$ represent the ground truth depth and predicted depth, respectively, $||\cdot||$ represents the SiLog distance.

\begin{table*}[!t]
	\setlength{\abovecaptionskip}{1pt}

	\caption{Quantitative results of the proposed method against the existing methods, using the Eigen split on the KITTI~\cite{geiger2013vision} dataset. '$ \downarrow$' indicates smaller is better, while '$\uparrow$' denotes larger is better. $\Delta$ RMSE is calculated by comparing with the baseline PixelFormer~\cite{agarwal2023attention}. { $^\dag$ refers to results obtained with both InvT-IndDiffusion and AV-LFE in compatible mode. $^\ddag$ refers to results obtained with both InvT-IndDiffusion and AV-LFE in fully trainable mode.} }
	\begin{center}
		\setlength{\tabcolsep}{1.5mm}
		
		\begin{tabular}{c l c c c c c c c c} \hline
			\toprule[0.8pt]
			Type &Method & RMSE $\downarrow$ &$\Delta$ RMSE $\uparrow$ & {Abs Rel $\downarrow$} &{Sq Rel $\downarrow$} &RMESlog $\downarrow$ &$\zeta_1\uparrow$ &$\zeta_2\uparrow$ &$\zeta_3\uparrow$ \\ \hline
			\multirow{13}*{Regression} & Eigen et al.~\cite{eigen2014depth} & 6.307 & -203.07\% & 0.203 & 1.548 & - & - & - & -  \\ 
			~ & 	P3Depth \cite{patil2022p3depth} & 2.842      & -36.56\% & 0.071 & 0.270 & 0.103 & 0.953 & 0.993 & {0.998}  \\ 
			~ & DORN~\cite{fu2018deep} & 2.727 & -31.04\% & 0.072 & 0.307 & 0.120 & 0.932 & 0.984 & 0.995  \\ 
			~ & NeWCRFs~\cite{yuan2022new} & 2.129 & -2.31\% & 0.052 & 0.155 & 0.079 & 0.974 & 0.997 & 0.999  \\ 
			~ & URCDC~\cite{URCDC} & 2.032 & 2.35\% & 0.050 & 0.142 & 0.076 & 0.977 & 0.997 & 0.999  \\ 
			~ & EVP~\cite{lavreniuk2023evp} & 2.015 & 3.17\% & 0.048 & 0.136 & 0.073 & 0.980 & 0.999 & 1.000  \\ 
			~ & NDDepth~\cite{shao2023nddepth} & 2.025 & 2.69\% & 0.050 & 0.141 & 0.075 & 0.978 & 0.998 & 0.999  \\ 
			~ & DDP~\cite{ji2023ddp} & 2.072 & 0.43\% & 0.050 & 0.148 & 0.076 & 0.975 & 0.997 & 0.999  \\ 
			~ & P3Depth~\cite{patil2022p3depth} & 2.842 & -36.57\% & 0.071 & 0.270 & 0.103 & 0.953 & 0.993 & 0.998  \\ 
			~ & Depthformer~\cite{li2023depthformer} & 2.143 & -2.98\% & 0.052 & 0.158 & 0.079 & 0.975 & 0.997 & 0.998  \\ 
			
			~ & ECoDepth~\cite{patni2024ecodepth} & 2.039 & 2.02\% & 0.048 & 0.139 & 0.074 & 0.979 & 0.997 & 1.000  \\ 
             ~ & {WorDepth~\cite{zeng2024wordepth}} & {2.039} & {2.02\%} & {0.049} & {-} & {0.074} & {0.979} & {0.998} & {0.999}  \\
            ~ & {DiffusionDepth~\cite{duan2024diffusiondepth}} & {2.016} & {3.12\%} & {0.050} & {0.141} & {0.074} & {0.977} & {0.998} & {0.999}  \\ \hline
			~& Adabin~\cite{bhat2021adabins} & 2.360 & -13.41\% & 0.058 & 0.190 & 0.088 & 0.964 & 0.995 & 0.999  \\ 
			~& BinsFormer~\cite{li2024binsformer} & 2.141 & -2.88\% & 0.053 & 0.156 & 0.080 & 0.974 & 0.997 & 0.999  \\ 
			\multirow{2}*{Classification-}& iDisc \cite{piccinelli2023idisc} & {2.067}       & 0.67\% & {0.050} & {0.145} & {0.077} & {0.977} & 0.997 & {0.999}  \\ 
			\multirow{2}*{Regression}  & GEDepth~\cite{yang2023gedepth} & 2.054 & 1.30\% & 0.049 & 0.143 & 0.076 & - & - & -  \\ 
			~ & PixelFormer~\cite{agarwal2023attention} & 2.081 & 0\% & 0.051 & 0.149 & 0.077 & 0.976 & 0.997 & 0.999  \\ 
			~& {IID-RDepth} & 1.996 & 4.09\% & 0.050 & 0.140 & 0.075 & 0.979 & 0.998 & 1.000  \\ 
			~& {IID-RDepth}$^\dag$ & 1.722 & 17.25\% & 0.047 & 0.107 & 0.069 & 0.984 & 0.999 & 1.000  \\ 
			~&{IID-RDepth}$^\ddag$ & \textbf{1.295} & \textbf{37.77\% }& \textbf{0.034} & \textbf{0.057} & \textbf{0.052} & \textbf{0.992} & \textbf{0.999} & \textbf{1.000}  \\ 
			\bottomrule[0.8pt]
		\end{tabular}
		\label{Tab:kitti}%
	\end{center}
	\vspace{-0.5cm}
\end{table*}

\subsection{Auxiliary Viewpoint based Low-level Feature Enhancement Module (AV-LFE)}

The above InvT-IndDiffusion restores the high-level {features}, which significantly affects the depth estimation performance. On the other hand, the low-level features can also help improve the depth estimation to some extent as shown in Fig.~\ref{fig:RMSE} in {the} Motivation. However, low-level features mostly provide local {detailed} features which cannot be appropriately processed by InvT-IndDiffusion.
Therefore, an AV-LFE ({Auxiliary Viewpoint based} Low-level Feature Enhancement Module) module is developed to improve the shortcut {connections} used in the UNet architecture, to enhance the low-level spatial features by introducing auxiliary viewpoints when available. { To be specific, multi-view images are available in many scenarios including some autonomous driving datasets~\cite{geiger2013vision}, and the views, other than the one used in the MDE, can serve as  auxiliary viewpoints.}

Given that there exist certain scenarios without auxiliary viewpoints, our AV-LFE module can be operated in two modes, i.e{.}, the compatible mode and fully trainable mode. For the compatible mode, AV-LFE is used as a plug-and-play module where the AV-LFE is trained and used while the parameters of the encoder and decoder are frozen. In this case, the AV-LFE is a selectable module which improves the performance with available auxiliary {viewpoint} and {shows} no effect without it. For the fully trainable mode, the whole network with the AV-LFE {module} is trained to achieve optimal performance.

Specifically, the low-level features of the auxiliary viewpoint are extracted by the first two layers of the same encoder used to process the main viewpoint. The proposed AV-LFE module takes both the encoder features of the main and auxiliary viewpoints as {inputs} and processes them {for} the decoder to enhance the original shortcut connection. The structure of the AV-LFE module is shown in Fig.~\ref{fig:frame}. The features of the auxiliary viewpoint are first aligned to the main viewpoint using the deformable convolution, in order to reduce the {disparities} between them. Then the features are fused through a convolution layer, to reduce the dimension to the same one as the encoder feature of the main viewpoint. {The formulation of the AV-LFE module is as follows:}
\begin{equation}
	\begin{aligned} 
		\label{ave}
		F_{align }=f_{DfConv }(f_{main}, f_{aux}) \\
		Output =f_{Conv2d}(F_{align})
	\end{aligned}
\end{equation}
where $f_{main}$ and $f_{aux}$ represent the features of the main view and auxiliary view, respectively. $f_{DfConv}$ is composed of a $3\times3$ deformable convolution with a GELU activation function for alignment. $f_{Conv}$ consists of two layers of $3\times3$ convolutions with GELU activation functions for  aligned {features} processing.

In this way, the whole network does not need to be changed and only the shortcut {connections} {are} replaced, thus providing compatibility with only using the main viewpoint.

\section{Experiments}
The widely used KITTI~\cite{geiger2013vision} and DDAD~\cite{packnet} datasets are adopted for evaluating the proposed method. The description of the datasets and the implementation details, including optimization settings and evaluation measures, are provided in the supplementary {materials} due to page limits.

\subsection{Experimental setting on the feature optimization}
{For the feature optimization experiment described in the Motivation, the parameters of the model (Pixelformer~\cite{agarwal2023attention}) pretrained on the KITTI dataset {are} frozen. {Per-image optimization {is} employed to refine encoder features at the skip connections, where the features {are} optimized {for} each image.} {Specifically, the network parameters are frozen and only the encoder features are optimized based on each image. It is used to indicate the potential of the network representation capability.} Features, at scales of 1/32, 1/16, 1/8, and 1/4, {are} independently optimized, thereby ensuring that adjustments in one layer {do not} impact others. The Adam optimizer was utilized, focusing on optimizing feature representations while keeping network parameters fixed. The experiment was conducted on the KITTI test dataset, which comprises 697 samples {with crops of 352×1216 resolution used in~\cite{garg2016unsupervised}}.} 


\subsection{Comparison with State-of the-Art Results}

\begin{figure*}[t]
	
	\centering
	\setlength{\abovecaptionskip}{5pt}
    \vspace{0pt}
	\includegraphics[width=1\linewidth]{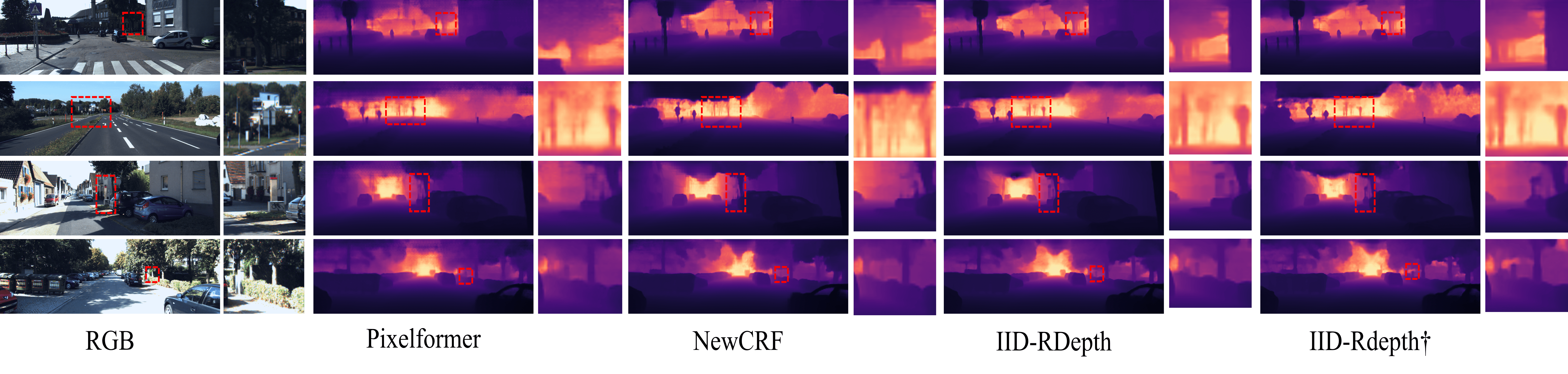}
    \vspace{-6pt}
	\caption{Qualitative results on the KITTI dataset comparing our IID-RDepth with existing methods.  IID-RDepth utilizes the proposed InvT-IndDiffusion, while IID-RDepth$^\dag$ combines both InvT-IndDiffusion and AV-LFE in a compatible mode. {With the proposed method, results of both fine-grained details and distant objects are improved. Some examples are zoomed such as  the gaps between trees in the first row, the geometric structure of distant road signs in the second row, the edges of nearby white stone monuments along the road in the third row,  the roadside signs in the fourth row.} }
	\label{fig:kitti}
\end{figure*}

\textbf{Results on KITTI:} The proposed IID-RDepth is compared with the state-of-the-art (SOTA) methods on the KITTI~\cite{geiger2013vision} benchmark, and the results are shown in Table~\ref{Tab:kitti}. It can be seen that the proposed method achieves comparable or superior performance to current SOTA methods. {In terms of RMSE reduction, the proposed IID-RDepth improves the performance by $4.08\%$ compared to PixelFormer. When the AV-LFE module is used, the performance is greatly improved. In the compatible mode, the RMSE is further reduced by $17.25\%$ compared to PixelFormer. In the fully trainable mode, it improves by $37.77\%$ compared to PixelFormer.}  {Due to the large areas of flat roads and sky contained in the KITTI dataset, accurate estimation {with} sparse ground truth labels {is} difficult. Moreover, our method focuses on general depth prediction without special processing of such areas. Thus, averaging performance improvements across the entire image may result in a relatively small improvement in terms of the Absolute Relative Error (Abs Rel). However, our method consistently improves the performance, which is further validated with a paired t-test shown in the ablation study.} Some example visual comparisons are shown in Fig.~\ref{fig:kitti}. It can be seen that our method more effectively models distant objects, resulting in clearer outlines. Depth estimation of distant objects {is} challenging due to their long distance and small size. The proposed InvT-IndDiffusion restores the high-level features to provide more clues {for} estimating the depth of distant objects. When utilizing the AV-LFE module, low-level features from the auxiliary viewpoint are integrated, allowing for the correction of {detailed} features from the main viewpoint and achieving richer details in the predicted depth map, as evidenced in Fig.~\ref{fig:kitti}.

\noindent\textbf{Results on DDAD:} We {evaluate} our method on {the} novel outdoor benchmark DDAD~\cite{packnet}{,} a relatively new dataset without many prior works reporting results. {Two baseline networks are tested and our method is incorporated into both baselines as a plug-and-play module. This can better validate our method as a feature restoration approach, {by} only using the proposed method to restore the features without changing the original overall architectures. To ensure fair comparison, only the {InvT-IndDiffusion} module is incorporated into the {baselines}.} The results are illustrated in Table~\ref{Tab:DDAD}. It can be seen that the proposed InvT-IndDiffusion consistently enhances network performance, validating the robustness and adaptability of our method across various network architectures and benchmark datasets. 

\begin{figure*}[t]
	\centering
	\setlength{\abovecaptionskip}{5pt}
    \vspace{0pt}
	\includegraphics[width=1\linewidth]{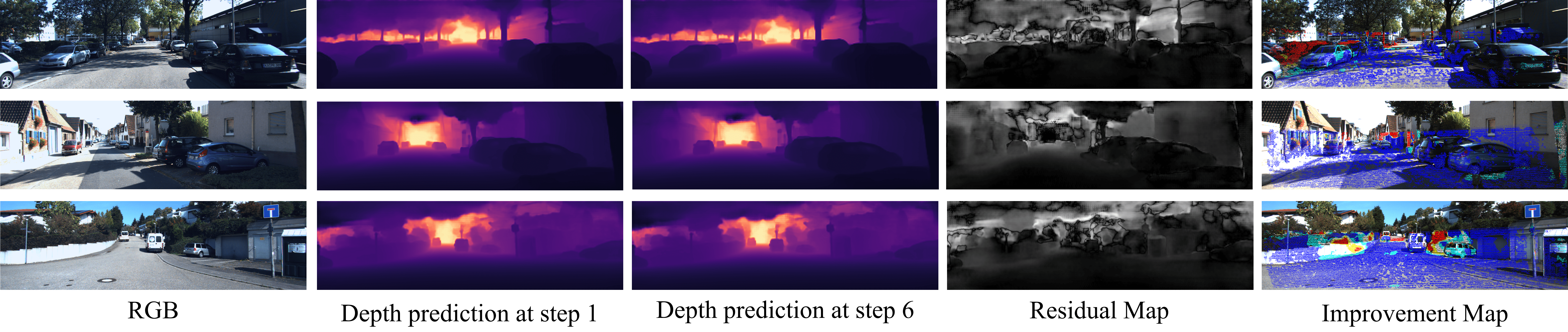}
    \vspace{-6pt}
	\caption{Visual illustration of the predicted depth maps and their differences when using the InvT-IndDiffusion with different interference steps, {where the blue to red colors represent improvement from small to large.} The residual map displays the differences of results obtained at different steps. {The improvement map further illustrates the reduction in prediction error towards the ground truth depth map, derived from the residuals between the results and the ground truth depth map at different steps, with blue to red indicating the magnitude of the improvement.}}
	\vspace{-0.3cm}
	\label{fig:erropmap}
\end{figure*}

\begin{table*}[t]
	\centering
    
	\caption{ Quantitative result comparison on the DDAD~\cite{packnet} dataset.  '$\downarrow$'indicates smaller is better, while '$\uparrow$' denotes larger is better. $\Delta$RMSE is calculated by comparing with the baseline PixelFormer~\cite{agarwal2023attention}. {ours-N (-P) means using NeWCRFs (PixelFormer) as the backbone.}}

	\begin{tabular}{c c c c c c c c c}
		
		\rowcolor{mycolor}
		\hline
		\multicolumn{9}{c}{\textbf{DDAD Benchmark}} \\ 
		\toprule[0.8pt]
		Method & RMSE$\downarrow$& $\Delta$RMSE$\uparrow$ & Abs Rel$\downarrow$ & Sq Rel$\downarrow$ & RMESlog$\downarrow$ & $\delta_1$ $\uparrow$ & $\delta_2$ $\uparrow$ & $\delta_3$ $\uparrow$  \\ \hline
		NeWCRFs~\cite{yuan2022new} & 5.21 & 9.23\% & 0.111 & 0.930 & 0.141 & 0.891 & 0.982 & \textbf{0.995}  \\ 
		PixelFormer~\cite{agarwal2023attention} & 5.74 & $\backslash$ & 0.127 & 0.134 & 0.157 & 0.861 & 0.972 & 0.993  \\ 
		Ours-P & 5.26 & 8.36\% & 0.121 & 0.988 & 0.146 & 0.881 & 0.977 & 0.994  \\ 
		Ours-N & \textbf{4.82} & \textbf{16.03\%} & \textbf{0.095} & \textbf{0.793} & \textbf{0.127} & \textbf{0.914} & \textbf{0.985} & \textbf{0.995}  \\ 
		\bottomrule[0.8pt]
	\end{tabular}
	\label{Tab:DDAD}
	\vspace{-0.2cm}
\end{table*}
\subsection{Ablation Study}
Ablation experiments on the InvT-IndDiffusion module, {the} invertible decoder{,} {the} number of parameters, and performance at different distances are conducted.

\noindent\textbf{Evaluation on the InvT-IndDiffusion module:} Ablation experiments are first conducted to demonstrate the effectiveness of the proposed InvT-IndDiffusion and explore the effect of various diffusion steps on the performance. The model without diffusion is tested{,} and for fair comparison, the invertible decoder is also used. The quantitative results are presented in Table~\ref{Tab:Diffusion}. It can be seen that our InvT-IndDiffusion improves the performance, with an RMSE reduction of $3.76\%$. Then the effect of the number of diffusion steps {is} tested and the quantitative results using only one step and six steps are shown in Table~\ref{Tab:Diffusion}, where the diffusion with six steps performs better. The qualitative results are shown in Fig.~\ref{fig:erropmap} along with the residual map and improvement map. The {residual} {maps} and improvement {maps} demonstrate that an increased number of diffusion steps effectively corrects and refines previously inaccurate estimates, particularly in regions such as edges and distant {scenes}.

\begin{figure*}[t]
	
	\centering
	\includegraphics[width=0.95\linewidth]{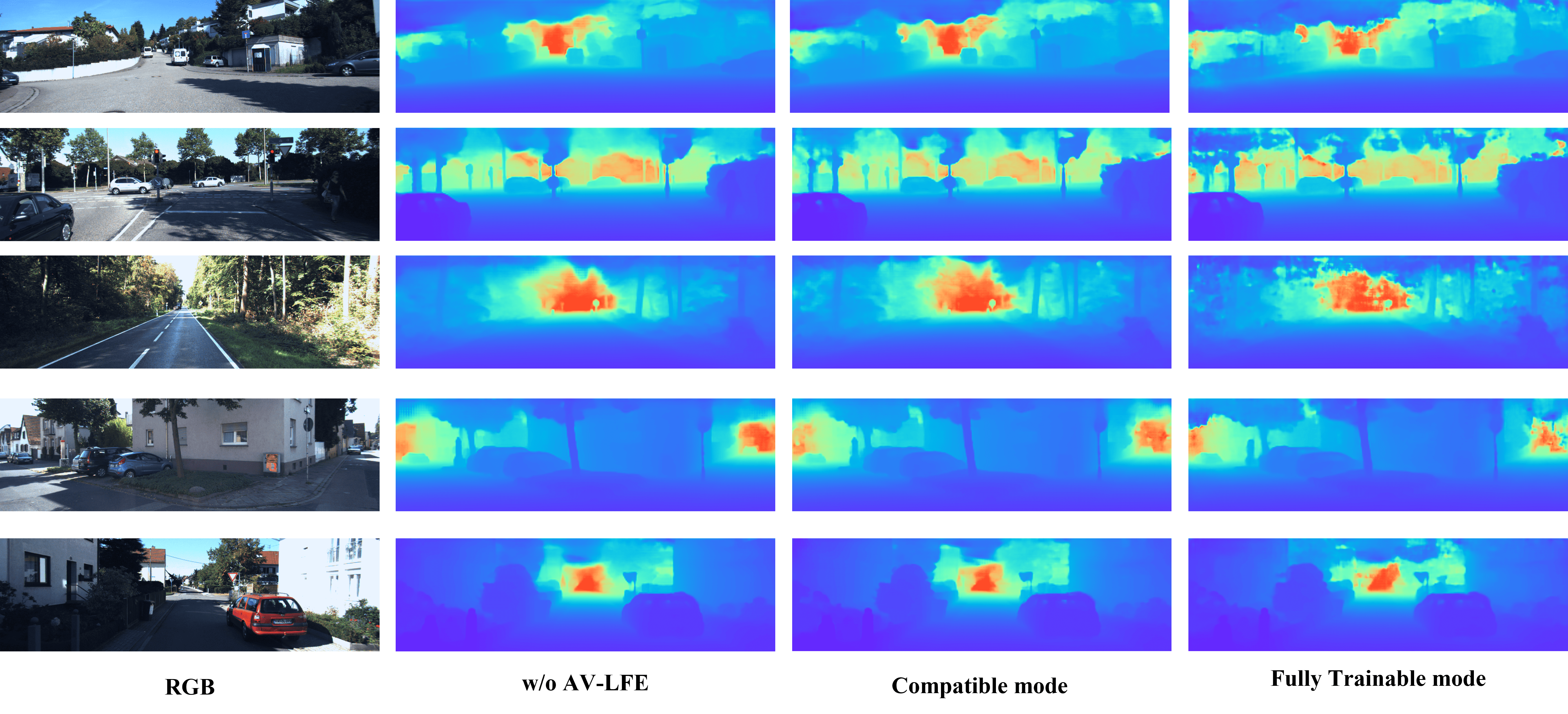}
   
	\caption{Qualitative Results of Ablation Studies on the AV-LFE Module in the KITTI~[48] Dataset. {A different pseudo color scheme, changing the colormap by {replacing the default Magma colormap to Jet}, is used to better illustrate the close-range depth result, with blue to red denoting increasing depth.}}
	\label{fig:auxsave}	
    \vspace{-0.6cm}
\end{figure*}

\begin{table}[t]
	\setlength{\tabcolsep}{1.5mm}
	\centering

	\caption{Ablation experiment on the InvT-IndDiffusion module. Diffusion-1 (-6) represents diffusion with 1 (6) inference steps. }

{	 
	\begin{tabular}{l c c  c c }
		\toprule[1pt]
		Method & RMSE$\downarrow$ & $\Delta$ RMSE$\uparrow$ &   Sq Rel$\downarrow$ & $\delta_1 \uparrow$ \\ \hline
		w/o Diffusoin  & 2.074 & - &   0.148 & 0.976 \\
		Diffusion-1 & 2.020 & 2.60\% &  0.142 & 0.978  \\ 
		Diffusion-6 & \textbf{1.996} &\textbf{3.76\%} &  \textbf{0.140} & \textbf{0.979}  \\ 
		\bottomrule[0.8pt]
	\end{tabular}
}
\label{Tab:Diffusion}
\vspace{-0.3cm}
\end{table}

\noindent\textbf{Evaluation on the invertible decoder:} Different decoders are tested with the diffusion, including Transformer-based (TF), Convolution-based (Conv), and Invertible-based decoders (Inv). {The} results are shown in Table~\ref{Tab:decoders}. Given the inherent constraints of INNs, {their} expressive capability is generally weaker compared to CNNs and transformers in other tasks~\cite{dinh2014nice}. However, for the indirect diffusion as in our work, the proposed invertible decoder {achieves} the best performance, validating our analysis {in} section \ref{bbb}.

\begin{table}[t]
	\centering

	\caption{ Results using different types of decoders for the restored features by diffusion.}

	{
		
		\begin{tabular}{c c c  c c}
			\toprule[1pt]
			
			Method & RMSE$\downarrow$& $\Delta$ RMSE$\uparrow$ & Sq Rel$\downarrow$ & $\delta_1 \uparrow$  \\ \hline
			TF & 2.054 & -   & 0.147 & 0.976  \\ 
			Conv & 2.017 & 1.80\% &  0.144 & 0.977  \\ 
			Inv & \textbf{1.996}& \textbf{2.82\%} &  \textbf{0.140} & \textbf{0.979}  \\ \bottomrule[0.8pt]
		\end{tabular}
		\label{Tab:decoders}
	}
\end{table}


\noindent\textbf{Evaluation on the AV-LFE module}: {The AV-LFE module is evaluated to demonstrate the effectiveness of using an auxiliary view when available. It is trained using the {KITTI} dataset, where both left and right viewpoints are available. The left view is used as the main view and the right {is used} as the auxiliary view.} The quantitative results {are} as shown in Table~\ref{Tab:kitti}. It can be seen that the AV-LFE module significantly enhances the performance. In {the} {Compatible/Fully Trainable} modes, the model achieves improvements of 17.25\%/37.77\% in terms of RMSE and 7.84\%/33.33\% in Abs Rel. The qualitative results are depicted in Fig.~\ref{fig:auxsave}. The pseudo-depth {maps} here employs a different pseudo color strategy, aiming to highlight the detailed variations in the close-range depth {results}. {It can be observed that the proposed module generates depth maps with sharper edges under {Fully Trainable and Compatible} modes (e.g., Row 1: power poles; Row 2: traffic lights). The {AV-LFE} module also shows higher sensitivity to fine details and subtle depth variations (e.g., Row 3: nearby trees; Row 4: stone monument; Row 5: close/distant road signs).}


\noindent\textbf{Evaluation on depth prediction at different distances}: To better understand the effectiveness of IID-RDepth, the depth at different ranges {is} evaluated separately, and the results are shown in Table~\ref{Tab:Range} with comparison to the baseline~\cite{agarwal2023attention}. It can be seen that while our method performs better {at} all distances, it performs significantly better {in long-range} depth prediction with a reduction of $8.05\%$ in terms of RMSE, and $14.46\%$ in terms of Sq Rel, {which is} attributed to InvT-IndDiffusion enhancing the perception of distant objects.
\begin{table}[t]
	\centering
 
	\caption{Results in terms of different depth ranges, including near (0-20m), medium (20-50m) and far (50-80m).}
    
	\setlength{\tabcolsep}{1mm}
	{
		\begin{tabular}{c c c c c c}
			\toprule[1pt]
			Methods & Range & RMSE$\downarrow$ & $\Delta$ RMSE$\uparrow$ &  Sq Rel$\downarrow$ & $\delta_1 \uparrow$  \\ \hline
			Pixelformer & 	\multirow{2}*{0-20m} & 0.803 & -  & 0.056 & 0.988  \\ 
			IID-RDepth & ~ & \textbf{0.781} & \textbf{2.74\%} & \textbf{0.053} & \textbf{0.989}  \\ 
			Pixelformer & \multirow{2}*{20-50m}& 3.514 & -  & 0.433 & 0.935  \\ 
			IID-RDepth & ~ & \textbf{3.477} &\textbf{2.41\%} &\textbf{ 0.428} &\textbf{ 0.938}  \\ 
			Pixelformer & \multirow{2}*{50-80m} & 9.128 & - & 1.528 & 0.828  \\ 
			IID-RDepth & ~ & \textbf{8.393} & \textbf{9.02\%} & \textbf{1.307} & \textbf{0.862}  \\ \bottomrule[0.8pt]
		\end{tabular}
	}
	\label{Tab:Range}
\end{table}

\noindent\textbf{Statistical evaluation on the performance improvement significance}: To further demonstrate the effectiveness of the proposed {InvT-IndDiffusion} Model (IID-RDepth), we employed statistical testing to assess the significance of the improvements. Specifically, a paired t-test is used with two {hypotheses}. The null hypothesis ($H_0$) posits that there is insufficient evidence to suggest that the error of our method is significantly lower than that of the compared methods{,} whereas the alternative hypothesis ($H_1$) asserted that the error of our method is significantly lower.

The results are shown in Table~\ref{t-test}. It can be seen that our method outperforms the baseline model ({PixelFormer}) in  $97.2\%$ of the images, with p-values less than $0.05$. This result remains consistent under the more stringent criterion of $p < 0.01$, with $96.2\%$ of the images meeting the condition. Therefore, we have compelling evidence to conclude that the {results} of IID-RDepth model is superior to that of the compared methods, allowing us to reject the null hypothesis ($H_0$) in favor of the alternative ($H_1$). Furthermore, in the comparison with another method, iDisc, 95.5\% of the images satisfied the $p < 0.01$ criterion, further demonstrating that its error is substantially lower than that of the other methods. 

\begin{table}[t]
	\centering
  
	\caption{{Paired t-test for improvement significance evaluation. The numerical values listed under each criterion indicate the number of pixels that our method outperforms the respective competitor.}}

	\begin{tabular}{c|c c c}
		\toprule[1pt]
		Method & Total number & $P<0.05$ & $P<0.01$ \\ \hline
		Ours / NewCRFs~\cite{yuan2022new} & 652 & 636 & 620 \\ 
		Ours / Pixelformer~\cite{agarwal2023attention} & 652 & 634 & 627 \\ 
		Ours / idisc~\cite{piccinelli2023idisc} & 652 & 632 & 623 \\ \hline
		
	\end{tabular}
    \vspace{-0.1cm}
	\label{t-test}
\end{table}

\section{Conclusion}
In this paper, we propose an IID-RDepth framework to solve the depth estimation task from the perspective of feature restoration. It is motivated {by} the observation that depth prediction performance can be significantly improved with better high-level features. An InvT-IndDiffusion is further developed for high-level feature restoration {via} diffusion. An invertible decoder is used to align the optimization direction of both the decoder and the diffusion model based on the bi-Lipschitz condition. In this way, InvT-IndDiffusion mitigates the feature deviations during the iterative optimization of the diffusion model when indirectly supervised with the final task loss instead of {explicit} feature supervision. Additionally, to enhance the low-level features, a plug-and-play AV-LFE module is designed to fully explore the available multi-view information. Experiments {demonstrate} that the proposed method achieves state-of-the-art results, verifying its effectiveness and generalizability.

{
    \small
    \bibliographystyle{IEEEtran}
    \bibliography{main_diff}
    
}

\end{document}